\documentclass[conference]{IEEEtran}
\IEEEoverridecommandlockouts
\usepackage{cite}
\usepackage{amsmath,amssymb,amsfonts}
\usepackage{algorithmic}
\usepackage{graphicx}
\usepackage{textcomp}
\usepackage{xcolor}
\usepackage{etoolbox}
\usepackage{graphicx}
\usepackage[skip=0.333\baselineskip]{caption}
\usepackage{subcaption}
\usepackage{multirow}
\usepackage{multicol}
\usepackage{hyperref}
\usepackage{tabularx}
\usepackage{float}

\def\BibTeX{{\rm B\kern-.05em{\sc i\kern-.025em b}\kern-.08em
    T\kern-.1667em\lower.7ex\hbox{E}\kern-.125emX}}

\title{VLM-Auto: VLM-based Autonomous Driving Assistant with Human-like Behavior and Understanding for Complex Road Scenes}

\makeatletter
\newcommand{\linebreakand}{%
  \end{@IEEEauthorhalign}
  \hfill\mbox{}\par
  \mbox{}\hfill\begin{@IEEEauthorhalign}
}
\makeatother

\author{\IEEEauthorblockN{Ziang Guo\textsuperscript{*}}
\IEEEauthorblockA{\textit{Intelligent Space Robotics Laboratory} \\
\textit{Skoltech}\\
Moscow, Russia \\
ziang.guo@skoltech.ru}
\and
\IEEEauthorblockN{Zakhar Yagudin\textsuperscript{*}}
\IEEEauthorblockA{\textit{Intelligent Space Robotics Laboratory} \\
\textit{Skoltech)}\\
Moscow, Russia\\
Zakhar.Yagudin@skoltech.ru}
\and

\IEEEauthorblockN{Artem Lykov\textsuperscript{*}}
\IEEEauthorblockA{\textit{Intelligent Space Robotics Laboratory} \\
\textit{Skoltech}\\
Moscow, Russia\\
artem.lykov@skoltech.ru}
\linebreakand
\IEEEauthorblockN{Mikhail Konenkov}
\IEEEauthorblockA{\textit{Intelligent Space Robotics Laboratory} \\
\textit{Skoltech}\\
Moscow, Russia\\
mikhail.konenkov@skoltech.ru}
\and
\IEEEauthorblockN{Dzmitry Tsetserukou}
\IEEEauthorblockA{\textit{Intelligent Space Robotics Laboratory} \\
\textit{Skoltech}\\
Moscow, Russia\\
d.tsetserukou@skoltech.ru}

\thanks{* These authors contributed equally to this work.}
}

\begin{document}

\maketitle
\thispagestyle{empty}
\pagestyle{empty}

\begin{abstract}

Recent research on Large Language Models for autonomous driving shows promise in planning and control. However, high computational demands and hallucinations still challenge accurate trajectory prediction and control signal generation. Deterministic algorithms offer reliability but lack adaptability to complex driving scenarios and struggle with context and uncertainty. To address this problem, we propose VLM-Auto, a novel autonomous driving assistant system to empower the autonomous vehicles with adjustable driving behaviors based on the understanding of road scenes. A pipeline involving the CARLA simulator and Robot Operating System 2 (ROS2) verifying the effectiveness of our system is presented, utilizing a single Nvidia 4090 24G GPU while exploiting the capacity of textual output of the Visual Language Model (VLM). Besides, we also contribute a dataset containing an image set and a corresponding prompt set for fine-tuning the VLM module of our system. In CARLA experiments, our system achieved $97.82\%$ average precision on 5 types of labels in our dataset. In the real-world driving dataset, our system achieved $96.97\%$ prediction accuracy in night scenes and gloomy scenes. Our VLM-Auto dataset will be released at \href{https://github.com/ZionGo6/VLM-Auto}{\textit{https://github.com/ZionGo6/VLM-Auto}}.

\end{abstract}

\section{Introduction}

\begin{figure}
  \centering
  \vspace{0.0cm}
  \includegraphics[scale=0.32]{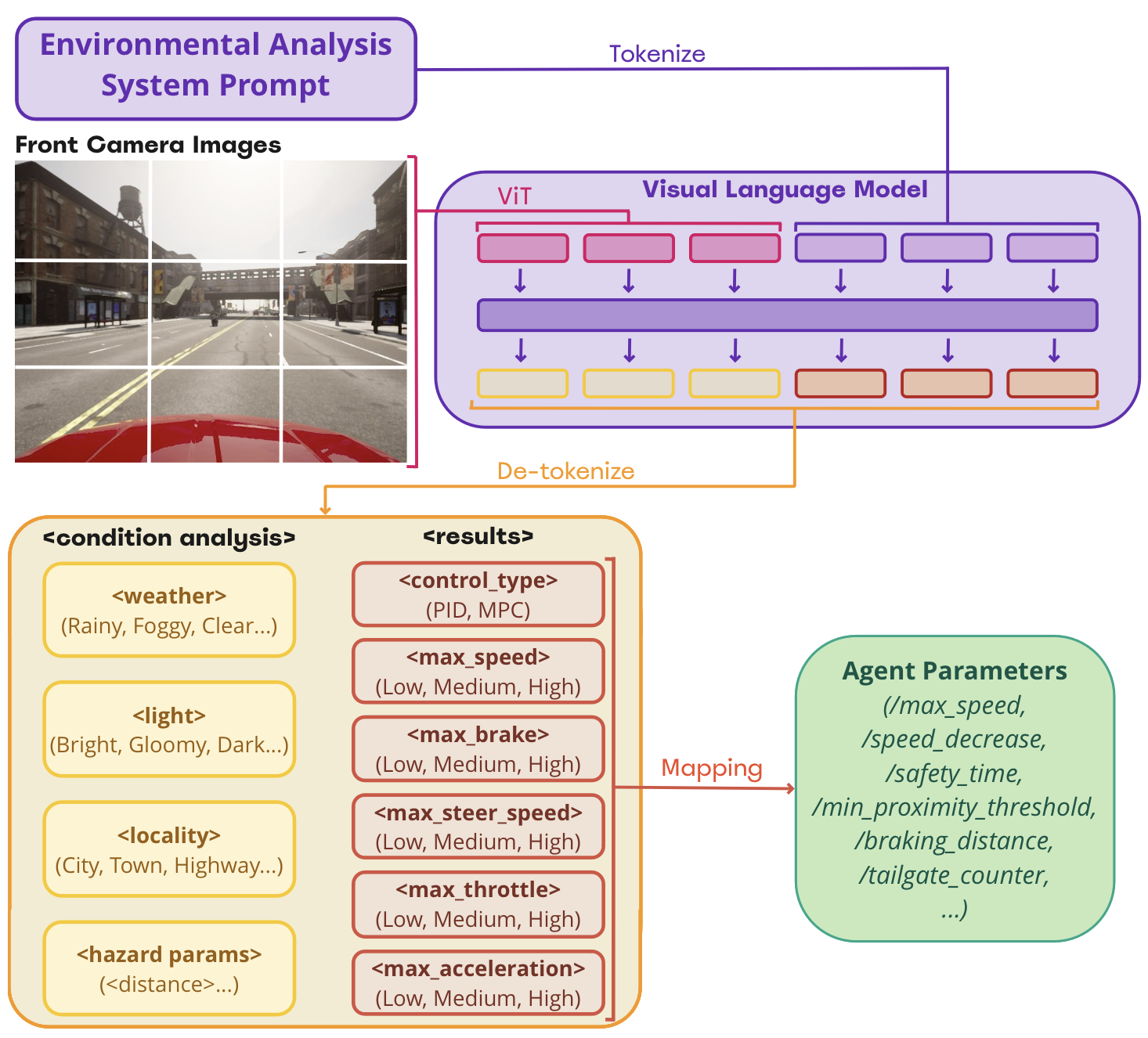}
  \caption{System layout of VLM-Auto. VLM module receives the image input and system prompt, publishing the analysis of environment and instruction results in a behavior tree format. Then the behavior tree of instruction results is mapped to the agent behaviors according to the analysis of the environment. }
  \label{fig:system_overview}
  \vspace{-0.2cm}
\end{figure}

\subsection{Motivation} 

In autonomous driving system development, two main solutions have been presented both in the academic and industrial fields till now. The first type is a modular system design with independent modules such as perception, prediction, control, planning, and etc. This design can empower distributed development and flexible extension, while the errors of such a system could accumulate because of asynchronization and delay of communication among modules \cite{hu2023planningoriented}. The second type is an end-to-end system design connecting sensor input and planning policy directly bypassing intermediate tasks and enabling the simple design of the network. However, end-to-end models demand their interpretability and logicality \cite{hu2022uniAD}. Generally, due to the complex traffic environment, long-tail data and dynamic scenes are still remaining limitations of these existing solutions \cite{shao2023lmdrive}. \par Promisingly, Large Language Models (LLMs) have been actively developed in recent years, bridging human interaction and reasoning \cite{gbagbe2024bivla}. Based on the advancements in LLMs, they can provide a more holistic understanding of the driving environment, allowing vehicles to respond more effectively to various driving scenarios with human-like logic which helps to alleviate public concerns about the safety and reliability of autonomous vehicles \cite{han2024dme}. However, to contribute to the precision and robustness requirements of autonomous driving, LLMs need more long-term verification and real-world experiments \cite{kambhampati2024llmplan}. \par In this work, we introduce VLM-Auto, an autonomous driving assistant to provide the instructions for driving behavior at an agent level based on the analysis of visual perception input.

\subsection{Related Work}

\subsubsection{End-to-End Autonomous Driving}

Recently, end-to-end autonomous driving has been developed vigorously. Shao et al. presented ReasonNet \cite{shao2023reasonnet}, an end-to-end driving framework that utilizes both temporal and global information of the driving scene to effectively predict the future evolution of the scene and behaviors of objects, especially in dense traffic scenarios. Jia et al. \cite{jia2023thinktwice} propose a cascaded decoder paradigm for predicting the future action of the ego vehicle in a coarse-to-fine fashion, inspired by the behavior of human drivers who check their intended target for safety and legitimacy. Hu et al. \cite{hu2022uniAD} propose a planning-oriented framework that incorporates full-stack driving tasks in one network, prioritizing planning and facilitating task coordination. Cui et al. \cite{cui2022coopernaut} introduce an end-to-end learning model that utilizes cross-vehicle perception for vision-based cooperative driving, aiming to enhance autonomous driving in dangerous or emergencies. Mao et al. \cite{mao2023gpt} developed an end-to-end model based on ChatGPT-3.5 for planning trajectory using command-prompt. Among them, ReasonNet approaches perception via global reasoning, but these methods still do not utilize and integrate a human-driver understanding of complex traffic scenes to fulfill the decision-making tasks.

\subsubsection{Large Language Models}

Nowadays, LLM building a strong foundation in Human-Robot interaction, enhancing communication ability of the robot. One of the recent fundamental works in applying LLM in robotics was conducted in Skoltech. Lykov et al. \cite{lykov2024cognitivedog} developed model and deployed it on the robotic dog to communicate and send control orders in natural language in real-time. Moreover, a recent study released the potential combination of large language models and autonomous driving. Li et al. \cite{li2024rule} present LLaDA, a tool that enables human drivers and autonomous vehicles to adapt their driving behavior to new locations by interpreting traffic rules using large language models. Sharan et al. \cite{sharan2023llm-assist} propose a hybrid planner, which combines a rule-based planner with an LLM-based planner. Shao et al. \cite{shao2023lmdrive} introduce a language-guided, end-to-end autonomous driving framework that integrates multimodal sensor data with natural language instructions, enabling interaction with humans and navigation software. Cui et al. \cite{cui2023talk2drive} introduce a framework that uses large language models to process verbal commands from humans and make autonomous driving decisions, taking into account contextual information and personalized preferences for safety, efficiency, and comfort. Wang et al. \cite{wang2023safety} explore the integration of Large Language Models (LLMs) into autonomous driving systems to enhance their performance and safety by leveraging the LLMs' common-sense knowledge, reasoning abilities, and human interaction capabilities. The above work mainly shows the exploitation of language modal and its extensions. However, in the autonomous driving field, a combination of multimodal sensors, especially including visual modal is critical for necessary scene understanding. 

\subsubsection{Visual Language Models in Autonomous Driving Scenarios}

In this section, we explore various approaches to integrating Visual Language Models (VLMs) into autonomous driving scenarios, highlighting their roles in environmental analysis and decision-making. DriveLM  \cite{sima2023drivelm} focuses on the integration of VLMs into end-to-end driving systems via Graph Visual Question Answering (VQA). By utilizing this approach, DriveLM enables comprehensive reasoning about driving scenes, encompassing perception, prediction, and planning stages. The introduced DriveLM-Data dataset and baseline approach provide a framework for end-to-end autonomous driving, showcasing competitive performance even when faced with unseen objects or sensor configurations. RAG-Driver \cite{yuan2024rag} addresses the crucial need for human-understandable explanations in autonomous driving systems. Employing retrieval-augmented multimodal large language models, RAG-Driver excels in producing detailed explanations for driving actions and accurate predictions of control signals. Its remarkable zero-shot generalization capabilities to previously unseen environments underscore its potential for real-world deployment. DriveVLM \cite{tian2024drivevlm} introduces an autonomous driving system that effectively leverages VLMs for enhanced scene understanding and planning. Through the integration of chain-of-thought (CoT) modules, DriveVLM achieves a robust spatial understanding and real-time inference speed. Particularly noteworthy is DriveVLM-Dual, a hybrid system that combines VLMs with traditional autonomous driving pipelines, resulting in superior performance in navigating complex and unpredictable driving conditions.\par All of the above research regarding VLMs needs heavy computation resources for both training and inference, which is a critical factor in the robustness and safety of autonomous driving. Besides, a hallucination of Large Language Models is still not explainable \cite{huang2023hallucination}, resulting in challenges and risks in practical deployment. For Large Language Models, to output the coordinates of waypoints with high precision and stable response for autonomous driving in complex traffic and extreme conditions needs more real-world experiments and long-term verification.

\subsection{Contribution}

Our main contributions to this work are as follows. An image dataset created in CARLA simulator \cite{Dosovitskiy17Carla} with defined weather, light, road surface, locality, and traffic conditions associated with a prompt dataset with control and behavior parameters based on the scenes defined in the image dataset. Besides, a pipeline of our VLM-Auto system is presented, where the CARLA simulator is used to run the simulation scenes, publishing the status information of the ego vehicle via Robot Operating System 2 (ROS2) \cite{ROS2} and our VLM module is wrapped within ROS2, reading the published ROS2 topics of front images of the camera on the ego vehicle. While analyzing the front images, our VLM module instructs and alters the driving behaviors of the ego vehicle in CARLA. Fig. \ref{fig:system_overview} shows our pipeline in detail, where a single Nvidia 4090 24G GPU is able to handle our whole pipeline.

\section{System Overview}

    Our system is driven by Qwen-VL by Bai et al. \cite{bai2023qwen}. Qwen-VL is a leading open-source model in the field of VLM, showcasing exceptional capabilities in tasks such as image captioning, question answering, visual localization, and interactive tasks. This model processes both textual and visual data and excels in recognizing individual objects and their positions, as well as grounding tasks, which are crucial for our study.
    
    Qwen-VL's high performance is attributed to its position-aware vision-language adapter and its efficient 3-stage training process. With a total of 9.6 billion parameters, the model comprises a visual encoder (1.9 billion parameters), a vision-language adapter (0.08 billion parameters), and the Qwen large language model (7.7 billion parameters).
    
    The advanced environmental analysis capabilities of Qwen-VL, combined with the reasoning power of Qwen \cite{bai2023qwentech}, make it particularly suitable for our task. The model's compact size allows for seamless integration into a self-driving car's onboard computer, enabling efficient local deployment without sacrificing performance. This positions Qwen-VL as an ideal choice for enhancing autonomous driving systems.
    
    System architecture is depicted in Fig. \ref{fig:system_overview}. The primary task of our system is to analyze the visual input from the front camera of the ego vehicle and draw conclusions about environmental information such as weather, light, road surface, locality, etc., and parameters of control such as maximum speed, maximum brake, maximum throttle, etc. Determining the driving behaviors of a self-driving car based on visual data is a complex task for VLMs. However, by breaking down the task into a two-step process, it becomes manageable. \par The task is decomposed to identify environmental information from an image by feeding specifically defined scenes from our image dataset to the model and to predict the levels of control and behavior parameters based on the described environmental data. Both of these tasks pose no significant challenges for a fine-tuned VLM, which ensures the practical pipeline of implementation in our proposed system. \par In the first step of the mentioned task, our VLM module receives system prompts containing the mission description and destination, along with the images from the ego vehicle's front camera. In this stage, the module identifies locality, lighting, and weather conditions, as well as potential hazards in front. Then our module continues to generate the levels of control and driving behavior parameters, guided by the environmental parameters identified in the first step. Lastly, all the obtained parameters are mapped as a set of agent behaviors altering and influencing the driving style of the ego vehicle in the CARLA simulator based on the image input of our VLM module. 

\section{Methodology}

\subsection{Dataset Collection}

\begin{figure}[t]
    \centering
    \includegraphics[scale=0.3]{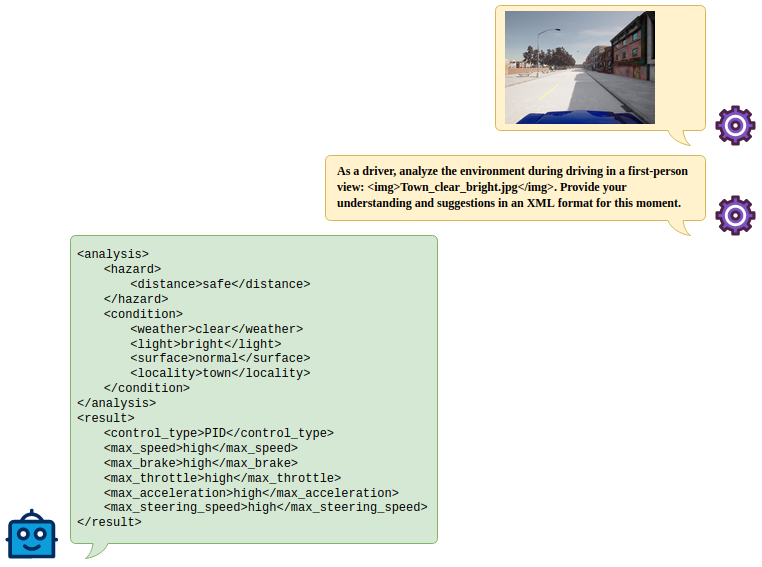}
    \caption{Example of our dataset with questioning and answering.}
    \label{fig:dataset_example}
\end{figure}

\par Our image dataset is collected in the CARLA simulator from the view of the front camera of ego vehicle under defined weather (clear, rainy, foggy), light (bright, gloomy, dark), locality (city, town, highway) conditions with a classification of safe and unsafe distance concerning the potential obstacle in front \cite{Dosovitskiy17Carla}. \par In our prompt dataset, system prompts are given as the request of accomplishment of the driving missions and notice of the environmental information from the perspective of a driver's view. Then we include the defined environmental information and the suggestions for vehicle control and driving behavior regarding control type, maximum speed, maximum brake, maximum throttle, maximum acceleration, and maximum steering speed as the output prompt in a behavior tree format. Here is an example of our dataset in Fig. \ref{fig:dataset_example}.

\subsection{Training Recipe}

The VLM of our system was trained on the foundation of the Qwen-VL architecture utilizing the Quantized Low-Rank Adaptation (QLoRA) method \cite{dettmers2024qlora}, which is a form of Parameter Efficient Fine-tuning (PEFT)\cite{fu2023peft}. During the training process, the weights of the visual encoder were kept frozen to focus on optimizing the language aspect of the model.

Training was carried out on a single Nvidia RTX 4090 GPU, which provided 24 GB of video memory for processing. The dataset, containing a total of 221,228 samples, was divided into batches of 6 samples each to maintain efficient training throughput. Additionally, gradient accumulation steps were set to 8, resulting in an epoch comprising approximately 4,600 steps.

With a learning rate of 1e-4, the model quickly adapted to the target emergent capabilities and responded to the desired format. This process only required one epoch of training, which took around 25 hours to complete. Despite the relatively short training time, the approach proved effective, yielding satisfactory results in terms of model performance and output quality.

\begin{figure}
    \centering
    \includegraphics[width=0.43\textwidth,height=\textheight, keepaspectratio]{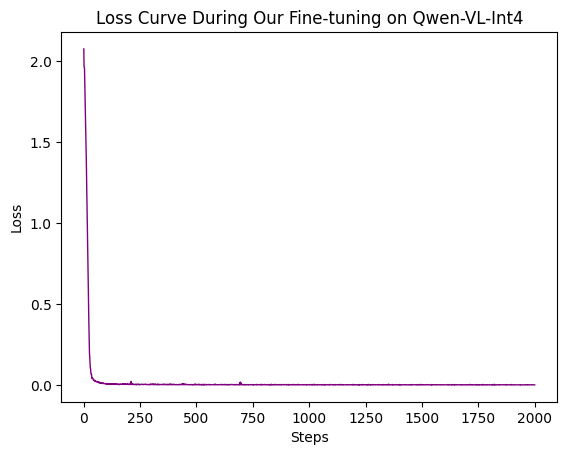}
    \caption{The training loss steadily decreased throughout the fine-tuning process and reached convergence after approximately 1,000 optimization steps.}
    \label{fig:loss_steps}
    \vspace{-1em}
\end{figure}

The progression of the training process is depicted in the training curve presented in Fig. \ref{fig:loss_steps}, showcasing the changes in loss over time and offering insights into the model's learning dynamics.

\section{Experimental Results}

\subsection{Experiment Setup}

To verify our system's effectiveness, we conducted two types of experiments. First, in CARLA, test scenes were created with adjustable weather, maps, and traffic settings. During the running of the test simulation, our VLM module was on, reading the front images from the ego vehicle and performing the scene understanding and behavior instructions. We recorded the driving scenes with vehicle trajectories and vehicle status information for verification. Second, we examined the generalization ability of our system's VLM module on HawkDrive dataset \cite{guo2024hawkdrive} with real-world driving scenes in adverse conditions. Based on the above experiments, we demonstrate our system's performance via the accuracy of our VLM module's prediction and the analysis of the ego vehicle status information.

\begin{figure}
    \centering
    \begin{subfigure}{.45\linewidth}
        \centering
        \includegraphics[width=\linewidth]{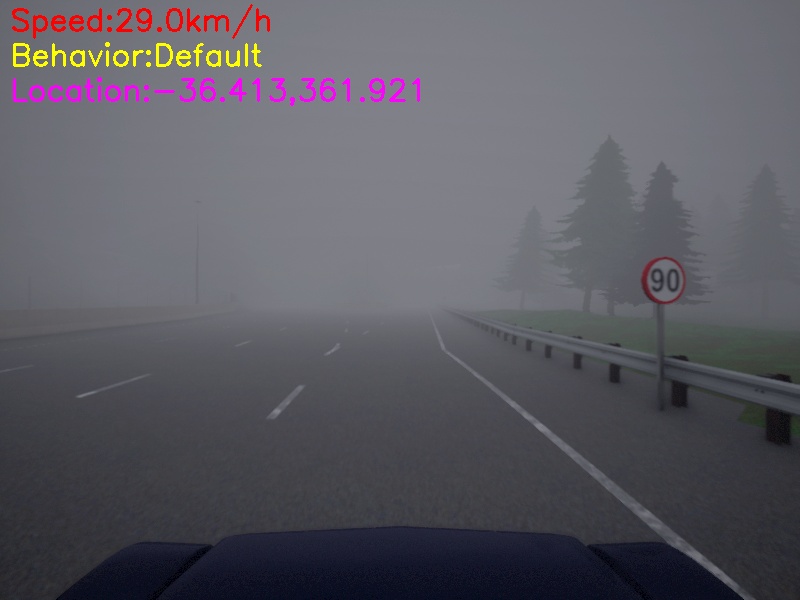}
    \end{subfigure}
    \begin{subfigure}{.49\linewidth}
        \centering
        \includegraphics[width=0.9\linewidth]{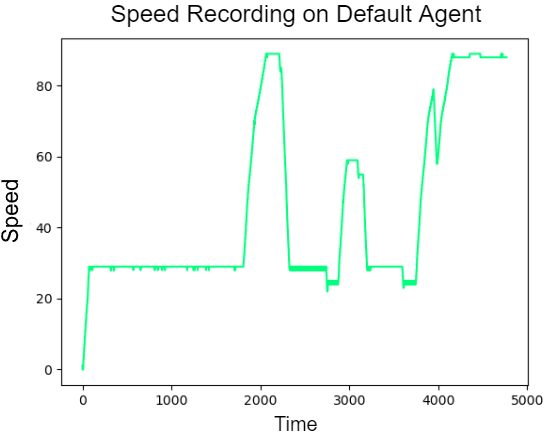}
    \end{subfigure}

    \begin{subfigure}{.45\linewidth}
    \vspace{0.5em}
        \centering
        \includegraphics[width=\linewidth]{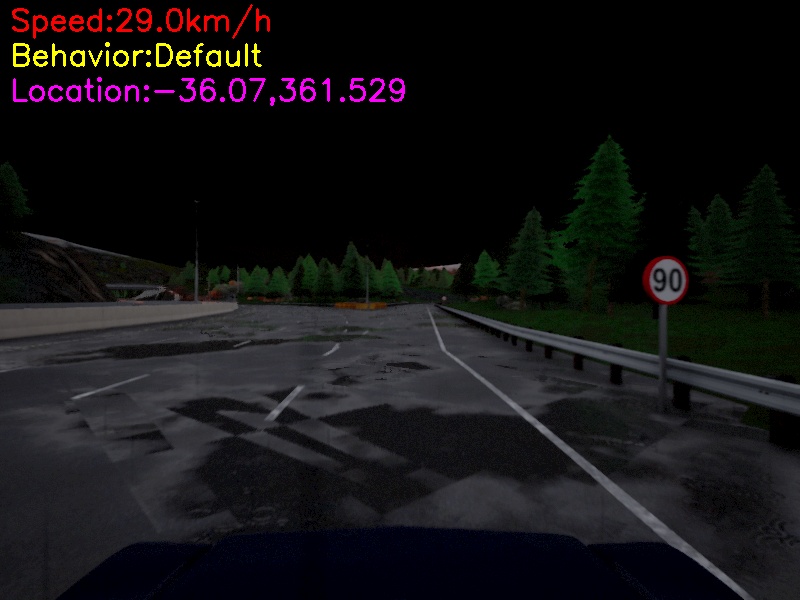}
        \caption{}
    \end{subfigure}
    \begin{subfigure}{.49\linewidth}
        \centering
        \includegraphics[width=0.9\linewidth]{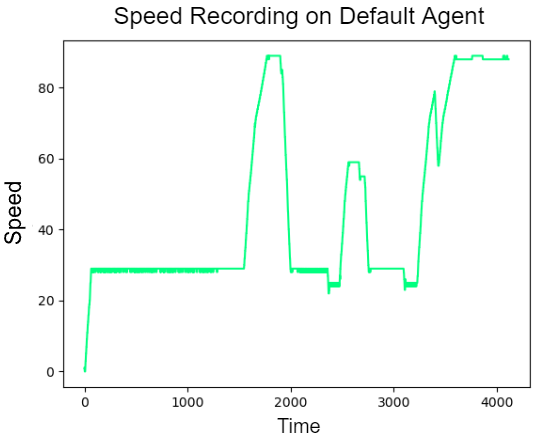}
        \caption{}
    \end{subfigure}

    \caption{Speed recording of the ego vehicle test with default built-in agent in CARLA map Town 04. a) The front images were recorded during the experiment in foggy and gloomy conditions (above) and rainy and gloomy conditions (below). b) The speed recording of the ego vehicle running with default built-in agent in foggy and gloomy conditions (above) and rainy and gloomy conditions (below).}
    \label{fig:default_agent}
\vspace{-1.5em}
\end{figure}

\subsection{CARLA Simulation}

In the CARLA simulator, we compared the driving behaviors between the default built-in agent and our VLM-Auto agent in Town 04, and Town 02. With the default agent, the ego vehicle was not able to switch driving behaviors along with the weather, light, and traffic conditions. In Fig. \ref{fig:default_agent}, under both rainy and foggy weather with gloomy light conditions, the driving behaviors of the ego vehicle along the same trajectory remained nearly identical since the planning and control modules were driven by defined rules only. When passing by the 90 km/h speed limit sign, the default agent guided the ego vehicle to reach 90 km/h, ignoring the adverse environmental information. 

\begin{figure}[H]
    \centering
    \begin{subfigure}{.45\linewidth}
        \centering
        \includegraphics[width=\linewidth]{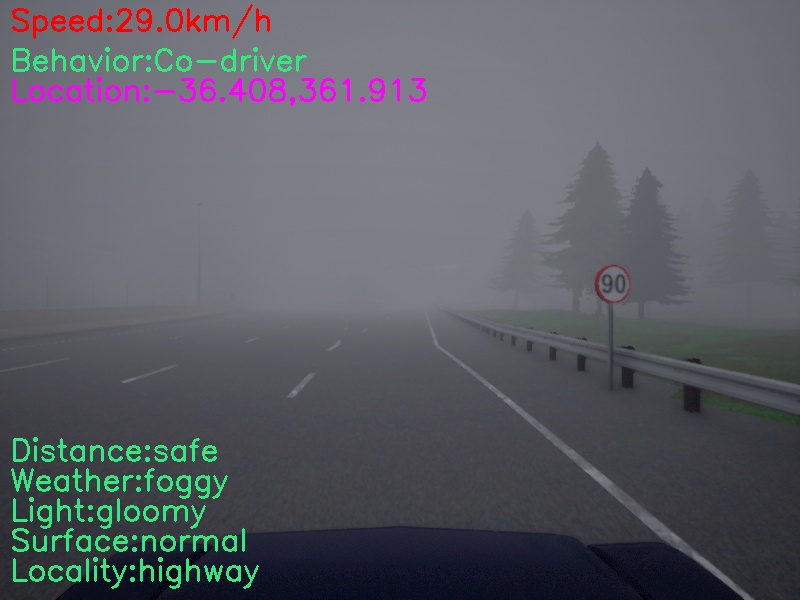}
    \end{subfigure}
    \begin{subfigure}{.49\linewidth}
        \centering
        \includegraphics[width=0.9\linewidth]{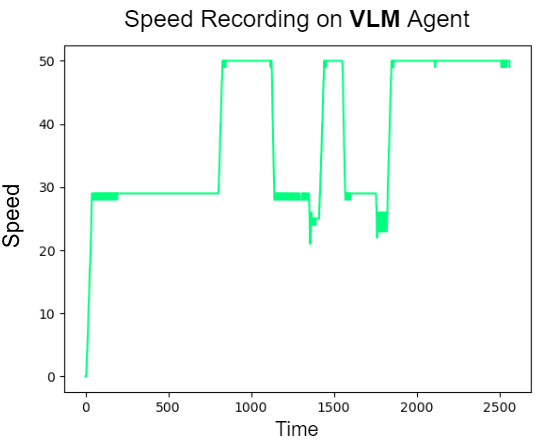}
    \end{subfigure}
    
    \begin{subfigure}{.45\linewidth}
    \vspace{0.5em}
        \centering
        \includegraphics[width=\linewidth]{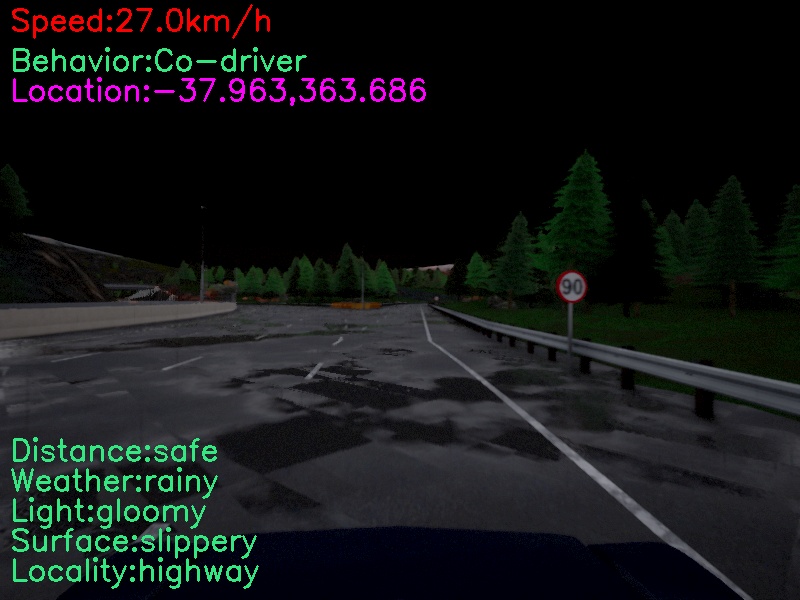}
        \caption{}
    \end{subfigure}
    \begin{subfigure}{.49\linewidth}
        \centering
        \includegraphics[width=0.9\linewidth]{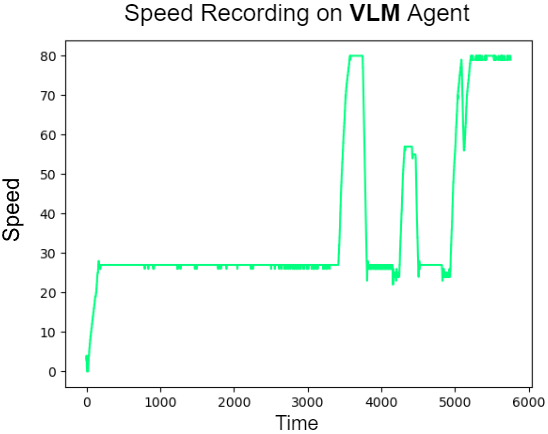}
        \caption{}
    \end{subfigure}

    \caption{Speed recording of the ego vehicle test with VLM-Auto agent in CARLA map Town 04. a) VLM-Auto agent running under foggy and gloomy conditions (above) and rainy and gloomy conditions (below). b) The speed of ego vehicle was adjusted according to the image input even if passing by the speed limit signs.}
    \label{fig:co-driver_agent}
\vspace{-0.50em}
\end{figure}

\begin{figure}
    \centering
    \begin{subfigure}{.45\linewidth}
        \centering
        \includegraphics[width=\linewidth]{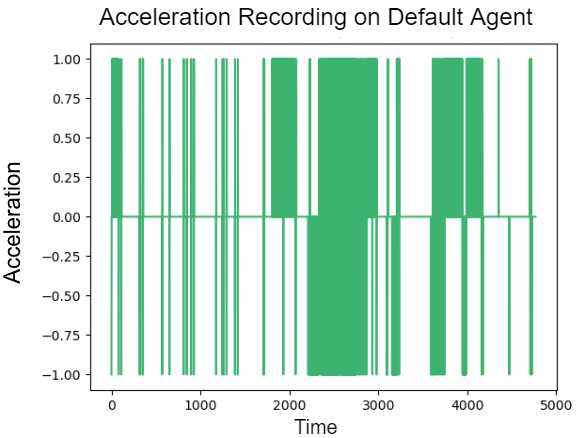}
    \end{subfigure}
    \begin{subfigure}{.49\linewidth}
        \centering
        \includegraphics[width=0.91\linewidth]{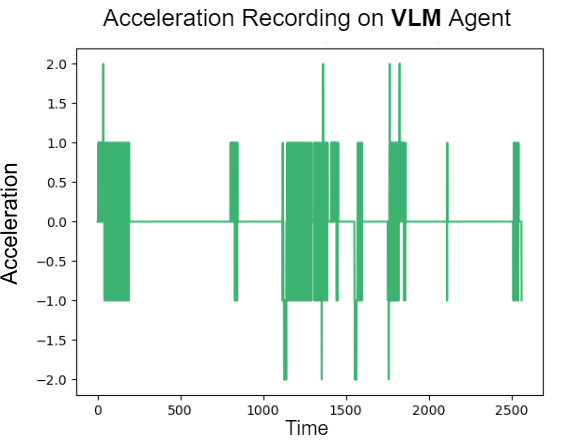}
    \end{subfigure}
    
    \begin{subfigure}{.45\linewidth}
    \vspace{0.5em}
        \centering
        \includegraphics[width=\linewidth]{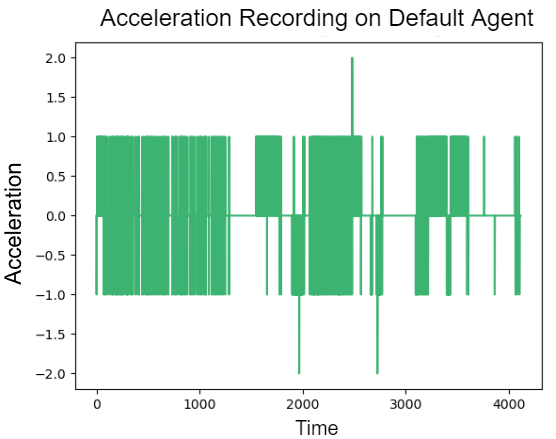}
        \caption{}
    \end{subfigure}
    \begin{subfigure}{.49\linewidth}
        \centering
        \includegraphics[width=0.92\linewidth]{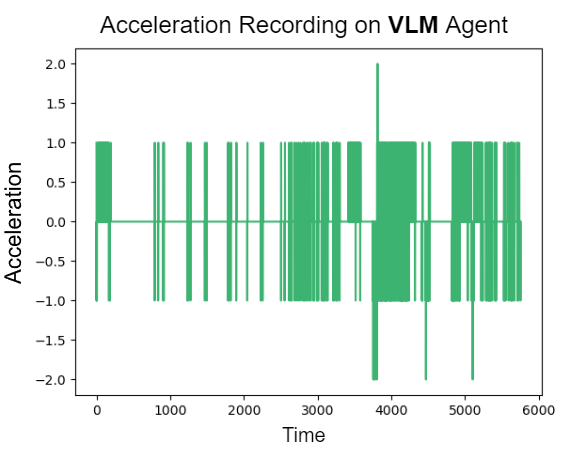}
        \caption{}
    \end{subfigure}

    \caption{Acceleration recording of the ego vehicle test with VLM-Auto agent in CARLA map Town 04. a) Acceleration recording of ego vehicle with default agent under foggy and gloomy conditions (above) and rainy and gloomy conditions (below). b) Acceleration recording of ego vehicle with VLM-Auto agent under foggy and gloomy conditions (above) and rainy and gloomy conditions (below).}
    \label{fig:acc}
\end{figure}

In Fig. \ref{fig:co-driver_agent}, with our VLM-Auto system running, under both rainy and foggy weather with gloomy light conditions, the driving behaviors of the ego vehicle were instructed according to the front image input. Based on the output of our VLM module, even if the ego vehicle passed by speed limit signs, our VLM-Auto system guided the ego vehicle to drive under the instructions since our VLM module performed analysis and understanding of the current driving scenes from the image input. Fig. \ref{fig:acc} displays the samples of the acceleration recording of our experiments. 

To interpret the results, we identified the relative maxima and minima throughout the acceleration recording. Then the frequency of fluctuations is calculated by counting the number of peaks and valleys in the data. Finally, we use the ratio of frequency of fluctuations and running time to denote the smoothness of driving behaviors as follows:

\begin{equation}
    \mathcal{\Dot{F}}_T = \frac{Concatenate(relmin(\mathcal{X}), relmax(\mathcal{X})) \times \frac{1}{2}}{T},
\end{equation}
where the arrays of indices of relative minima and relative maxima are concatenated as a combined array that contains the indices where the values in the input data $\mathcal{X}$ reach relative minima and maxima. Smaller $\mathcal{\Dot{F}}_T$ means smoother driving with less intensive fluctuation of acceleration. $T$ is the running time of our experiments. The comparison of acceleration recording is presented in Tab. \ref{acc}.

\begin{table}[H]
\begin{center}
\caption{Comparison of the smoothness of driving behaviors between default agent and VLM-Auto agent in Town 04 and Town 02.}
\begin{tabular}{c|cccc}
\hline
Conditions & \begin{tabular}[c]{@{}c@{}}Town 02\\ VLM-Auto\end{tabular} & \begin{tabular}[c]{@{}c@{}}Town 02\\ Default\end{tabular} & \begin{tabular}[c]{@{}c@{}}Town 04\\ VLM-Auto\end{tabular} & \begin{tabular}[c]{@{}c@{}}Town 04\\ Default\end{tabular} \\ \hline
Rainy \& Gloomy     & \textbf{0.043}                                                & 0.055                                                    & \textbf{0.104}                                                & 0.153                                                    \\ \hline
Foggy \& Gloomy     & \textbf{0.044}                                                & 0.045                                                    & \textbf{0.021}                                                & 0.117                                                    \\ \hline
Clear \& Bright     & \textbf{0.023}                                                & 0.029                                                    & \textbf{0.010}                                                & 0.019                                                    \\ \hline
\end{tabular}
\label{acc}
\end{center}
\end{table}

\par Tab. \ref{fig:town04_pred} displays the prediction accuracy of the Visual Language
Model module of our VLM-Auto system in CARLA map Town 04 under foggy gloomy condition and Town 02 under clear dark condition is presented, where VLM-Auto agent achieved $97.82\%$ average precision on 5 types of labels in our dataset.

\begin{table}[H]
\centering
\caption{The average prediction accuracy of the VLM module of our VLM-Auto system in CARLA map Town 04 under foggy gloomy condition and Town 02 under clear dark condition.}
\resizebox{120pt}{!}{\begin{tabular}{c|c}
\hline
\textbf{Categories} & \begin{tabular}[c]{@{}c@{}}\textbf{Accuracy}\end{tabular} \\ \hline
Distance     & 97.47\%\\ \hline
Weather     & 92.94\% \\ \hline
Light     & 99.8\% \\ \hline
Surface     & 98.8\% \\ \hline
Locality     & 99.7\% \\ \hline
\end{tabular}}
\label{fig:town04_pred}
\end{table}



\subsection{Real-World Driving Dataset}

\begin{figure}[H]
    \centering
    \begin{subfigure}{.45\linewidth}
        \centering
        \includegraphics[width=\linewidth]{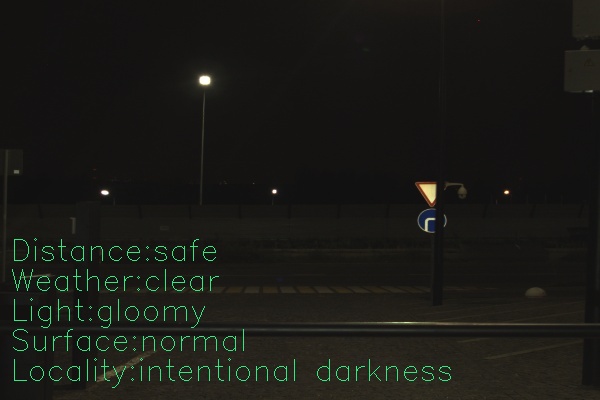}
    \end{subfigure}
    \begin{subfigure}{.45\linewidth}
        \centering
        \includegraphics[width=\linewidth]{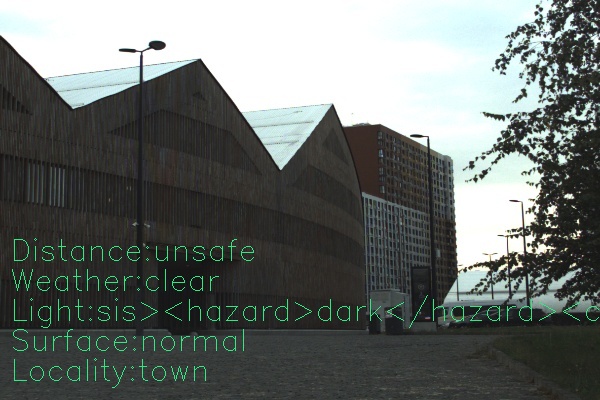}
    \end{subfigure}
    \begin{subfigure}{.45\linewidth}
    \vspace{0.5em}
        \centering
        \includegraphics[width=\linewidth]{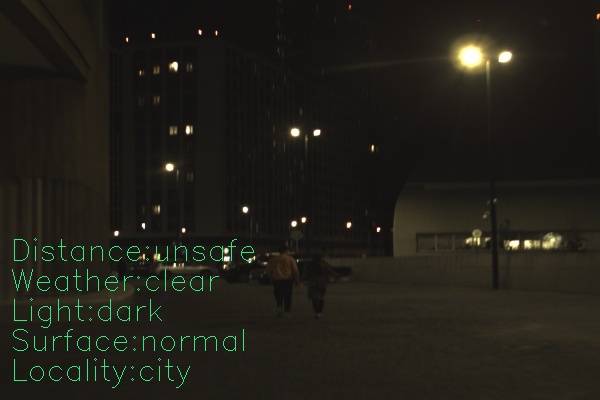}
        \caption{}
    \end{subfigure}    
    \begin{subfigure}{.45\linewidth}
        \centering
        \includegraphics[width=\linewidth]{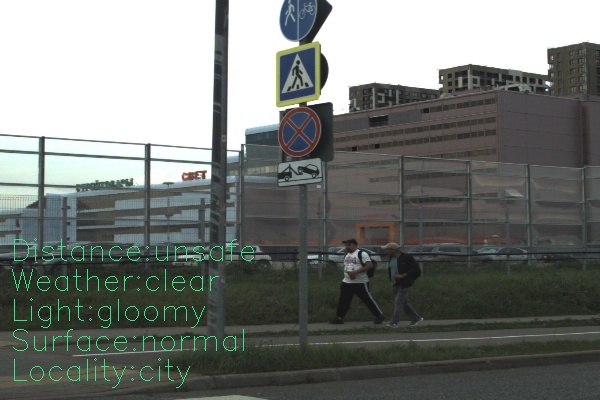}
        \caption{}
    \end{subfigure}
    \caption{VLM module of our VLM-Auto system test on HawkDrive dataset in the a) night scene. b) low light scene.}
    \label{fig:night_hawkdrive}
\vspace{-1.5em}
\end{figure}

To present the generalization ability of our system, HawkDrive dataset \cite{guo2024hawkdrive} which provides continuous driving scenes with different light conditions in a closed loop, is used to test the VLM module of our VLM-Auto system. In the adverse scenes of the dataset, we labeled the images corresponding to the ground truth regarding safety distance, weather, light condition, road surface and locality. The samples of the predictions are displayed in Fig. \ref{fig:night_hawkdrive}.

\begin{table}[H]
\begin{center}
\caption{The average prediction accuracy of the VLM module of our VLM-Auto system in the low light and night scenes.}
\resizebox{120pt}{!}{\begin{tabular}{c|c}
\hline
\textbf{Categories} & \begin{tabular}[c]{@{}c@{}}\textbf{Accuracy}\end{tabular} \\ \hline
Distance     & 98.67\%\\ \hline
Weather     & 98.61\% \\ \hline
Light     & 97.08\% \\ \hline
Surface     & 97.81\% \\ \hline
Locality     & 92.69\% \\ \hline
\end{tabular}}
\label{fig:night_pred}
\end{center}
\end{table}

Among 1,067 frames, prediction accuracy of each label is exhibited in Tab. \ref{fig:night_pred} in the low light and night scenes, showing a $96.97\%$ average precision on all the labels. \par To demonstrate our proposed system's capabilities, failure cases stemming from the hallucination output of the VLM module are presented in Fig. \ref{fig:night_hawkdrive}. For practical deployment, a safety guard mechanism can be implemented to filter out such hallucination outputs, thereby improving the accuracy of the VLM module's predictions.

\section{Conclusion}

In this work, we propose VLM-Auto, an autonomous driving assistant system to empower autonomous driving vehicles with adjustable driving behaviors based on the understanding of complex road scenes including safety distance, weather, light conditions, road surface and locality. To be practical, we present our system in a pipeline involving the CARLA simulator and Robot Operating System 2 (ROS2), while verifying the effectiveness of our system by comparing the driving behaviors of the rule-based default agent with our VLM-Auto agent. In CARLA experiments, our system achieved $97.82\%$ average precision on 5 types of labels in our dataset. In the real-world driving dataset, our system achieved $96.97\%$ prediction accuracy in night scenes and gloomy scenes. Besides, we also contributed a VLM-Auto dataset containing 221,228 image samples and a corresponding prompt set to spark further related research. Moreover, in terms of control smoothness we can see the impact of behavioral strategy in autonomous driving that could possibly minimize destabilization of controlled vehicle and enhance stability. \par From our results, the promising capacity of our VLM-Auto system is displayed. We also address several concerns encountered during this work. Specifically, the CARLA simulator was utilized for dataset collection due to its flexibility in creating customized traffic scenarios and its ability to label with natural languages. The generalization gap between simulation images and real-world images was reduced through fine-tuning with our VLM module. To further mitigate this gap, future efforts should focus on creating a consistent dataset that encompasses the necessary traffic scenarios. With the rapid advancement of Large Multimodal Models, our work provides insights that can facilitate further progress in the field of autonomous driving.

\section*{Acknowledgements} 
Research reported in this publication was financially supported by the RSF grant No. 24-41-02039.

\bibliographystyle{IEEEtran}
\bibliography{references}

\begin{thebibliography}{10}
\providecommand{\url}[1]{#1}
\csname url@samestyle\endcsname
\providecommand{\newblock}{\relax}
\providecommand{\bibinfo}[2]{#2}
\providecommand{\BIBentrySTDinterwordspacing}{\spaceskip=0pt\relax}
\providecommand{\BIBentryALTinterwordstretchfactor}{4}
\providecommand{\BIBentryALTinterwordspacing}{\spaceskip=\fontdimen2\font plus
\BIBentryALTinterwordstretchfactor\fontdimen3\font minus \fontdimen4\font\relax}
\providecommand{\BIBforeignlanguage}[2]{{%
\expandafter\ifx\csname l@#1\endcsname\relax
\typeout{** WARNING: IEEEtran.bst: No hyphenation pattern has been}%
\typeout{** loaded for the language `#1'. Using the pattern for}%
\typeout{** the default language instead.}%
\else
\language=\csname l@#1\endcsname
\fi
#2}}
\providecommand{\BIBdecl}{\relax}
\BIBdecl

\bibitem{hu2023planningoriented}
Y.~Hu, J.~Yang, L.~Chen, K.~Li, C.~Sima, X.~Zhu, S.~Chai, S.~Du, T.~Lin, W.~Wang, L.~Lu, X.~Jia, Q.~Liu, J.~Dai, Y.~Qiao, and H.~Li, ``Planning-oriented autonomous driving,'' 2023.

\bibitem{hu2022uniAD}
Y.~Hu, J.~Yang, L.~Chen, K.~Li, C.~Sima, X.~Zhu, S.~Chai, S.~Du, T.~Lin, W.~Wang \emph{et~al.}, ``Goal-oriented autonomous driving,'' \emph{arXiv preprint arXiv:2212.10156}, 2022.

\bibitem{shao2023lmdrive}
H.~Shao, Y.~Hu, L.~Wang, S.~L. Waslander, Y.~Liu, and H.~Li, ``Lmdrive: Closed-loop end-to-end driving with large language models,'' \emph{arXiv preprint arXiv:2312.07488}, 2023.

\bibitem{gbagbe2024bivla}
K.~F. Gbagbe, M.~A. Cabrera, A.~Alabbas, O.~Alyunes, A.~Lykov, and D.~Tsetserukou, ``Bi-vla: Vision-language-action model-based system for bimanual robotic dexterous manipulations,'' \emph{arXiv preprint arXiv:2405.06039}, 2024.

\bibitem{han2024dme}
W.~Han, D.~Guo, C.-Z. Xu, and J.~Shen, ``Dme-driver: Integrating human decision logic and 3d scene perception in autonomous driving,'' \emph{arXiv preprint arXiv:2401.03641}, 2024.

\bibitem{kambhampati2024llmplan}
S.~Kambhampati, K.~Valmeekam, L.~Guan, K.~Stechly, M.~Verma, S.~Bhambri, L.~Saldyt, and A.~Murthy, ``Llms can't plan, but can help planning in llm-modulo frameworks,'' \emph{arXiv preprint arXiv:2402.01817}, 2024.

\bibitem{shao2023reasonnet}
H.~Shao, L.~Wang, R.~Chen, S.~L. Waslander, H.~Li, and Y.~Liu, ``Reasonnet: End-to-end driving with temporal and global reasoning,'' in \emph{Proceedings of the IEEE/CVF Conference on Computer Vision and Pattern Recognition}, 2023, pp. 13\,723--13\,733.

\bibitem{jia2023thinktwice}
X.~Jia, P.~Wu, L.~Chen, J.~Xie, C.~He, J.~Yan, and H.~Li, ``Think twice before driving: Towards scalable decoders for end-to-end autonomous driving,'' in \emph{Proceedings of the IEEE/CVF Conference on Computer Vision and Pattern Recognition}, 2023, pp. 21\,983--21\,994.

\bibitem{cui2022coopernaut}
J.~Cui, H.~Qiu, D.~Chen, P.~Stone, and Y.~Zhu, ``Coopernaut: End-to-end driving with cooperative perception for networked vehicles,'' in \emph{Proceedings of the IEEE/CVF Conference on Computer Vision and Pattern Recognition}, 2022, pp. 17\,252--17\,262.

\bibitem{mao2023gpt}
J.~Mao, Y.~Qian, H.~Zhao, and Y.~Wang, ``Gpt-driver: Learning to drive with gpt,'' \emph{arXiv preprint arXiv:2310.01415}, 2023.

\bibitem{lykov2024cognitivedog}
A.~Lykov, M.~Litvinov, M.~Konenkov, R.~Prochii, N.~Burtsev, A.~A. Abdulkarim, A.~Bazhenov, V.~Berman, and D.~Tsetserukou, ``Cognitivedog: Large multimodal model based system to translate vision and language into action of quadruped robot,'' in \emph{Companion of the 2024 ACM/IEEE International Conference on Human-Robot Interaction}, 2024, pp. 712--716.

\bibitem{li2024rule}
B.~Li, Y.~Wang, J.~Mao, B.~Ivanovic, S.~Veer, K.~Leung, and M.~Pavone, ``Driving everywhere with large language model policy adaptation,'' \emph{arXiv preprint arXiv:2402.05932}, 2024.

\bibitem{sharan2023llm-assist}
S.~Sharan, F.~Pittaluga, M.~Chandraker \emph{et~al.}, ``Llm-assist: Enhancing closed-loop planning with language-based reasoning,'' \emph{arXiv preprint arXiv:2401.00125}, 2023.

\bibitem{cui2023talk2drive}
C.~Cui, Z.~Yang, Y.~Zhou, Y.~Ma, J.~Lu, and Z.~Wang, ``Large language models for autonomous driving: Real-world experiments,'' \emph{arXiv preprint arXiv:2312.09397}, 2023.

\bibitem{wang2023safety}
Y.~Wang, R.~Jiao, C.~Lang, S.~S. Zhan, C.~Huang, Z.~Wang, Z.~Yang, and Q.~Zhu, ``Empowering autonomous driving with large language models: A safety perspective,'' \emph{arXiv preprint arXiv:2312.00812}, 2023.

\bibitem{sima2023drivelm}
C.~Sima, K.~Renz, K.~Chitta, L.~Chen, H.~Zhang, C.~Xie, P.~Luo, A.~Geiger, and H.~Li, ``Drivelm: Driving with graph visual question answering,'' \emph{arXiv preprint arXiv:2312.14150}, 2023.

\bibitem{yuan2024rag}
J.~Yuan, S.~Sun, D.~Omeiza, B.~Zhao, P.~Newman, L.~Kunze, and M.~Gadd, ``Rag-driver: Generalisable driving explanations with retrieval-augmented in-context learning in multi-modal large language model,'' \emph{arXiv preprint arXiv:2402.10828}, 2024.

\bibitem{tian2024drivevlm}
X.~Tian, J.~Gu, B.~Li, Y.~Liu, C.~Hu, Y.~Wang, K.~Zhan, P.~Jia, X.~Lang, and H.~Zhao, ``Drivevlm: The convergence of autonomous driving and large vision-language models,'' \emph{arXiv preprint arXiv:2402.12289}, 2024.

\bibitem{huang2023hallucination}
L.~Huang, W.~Yu, W.~Ma, W.~Zhong, Z.~Feng, H.~Wang, Q.~Chen, W.~Peng, X.~Feng, B.~Qin \emph{et~al.}, ``A survey on hallucination in large language models: Principles, taxonomy, challenges, and open questions,'' \emph{arXiv preprint arXiv:2311.05232}, 2023.

\bibitem{Dosovitskiy17Carla}
A.~Dosovitskiy, G.~Ros, F.~Codevilla, A.~Lopez, and V.~Koltun, ``{CARLA}: {An} open urban driving simulator,'' in \emph{Proceedings of the 1st Annual Conference on Robot Learning}, 2017, pp. 1--16.

\bibitem{ROS2}
\BIBentryALTinterwordspacing
S.~Macenski, T.~Foote, B.~Gerkey, C.~Lalancette, and W.~Woodall, ``Robot operating system 2: Design, architecture, and uses in the wild,'' \emph{Science Robotics}, vol.~7, no.~66, p. eabm6074, 2022. [Online]. Available: \url{https://www.science.org/doi/abs/10.1126/scirobotics.abm6074}
\BIBentrySTDinterwordspacing

\bibitem{bai2023qwen}
J.~Bai, S.~Bai, S.~Yang, S.~Wang, S.~Tan, P.~Wang, J.~Lin, C.~Zhou, and J.~Zhou, ``Qwen-vl: A frontier large vision-language model with versatile abilities,'' \emph{arXiv preprint arXiv:2308.12966}, 2023.

\bibitem{bai2023qwentech}
J.~Bai, S.~Bai, Y.~Chu, Z.~Cui, K.~Dang, X.~Deng, Y.~Fan, W.~Ge, Y.~Han, F.~Huang \emph{et~al.}, ``Qwen technical report,'' \emph{arXiv preprint arXiv:2309.16609}, 2023.

\bibitem{dettmers2024qlora}
T.~Dettmers, A.~Pagnoni, A.~Holtzman, and L.~Zettlemoyer, ``Qlora: Efficient finetuning of quantized llms,'' \emph{Advances in Neural Information Processing Systems}, vol.~36, 2024.

\bibitem{fu2023peft}
Z.~Fu, H.~Yang, A.~M.-C. So, W.~Lam, L.~Bing, and N.~Collier, ``On the effectiveness of parameter-efficient fine-tuning,'' in \emph{Proceedings of the AAAI Conference on Artificial Intelligence}, vol.~37, no.~11, 2023, pp. 12\,799--12\,807.

\bibitem{guo2024hawkdrive}
Z.~Guo, S.~Perminov, M.~Konenkov, and D.~Tsetserukou, ``Hawkdrive: A transformer-driven visual perception system for autonomous driving in night scene,'' \emph{arXiv preprint arXiv:2404.04653}, 2024.

\end{thebibliography}

\end{document}